\documentclass[10pt,journal,letterpaper,twoside,compsoc]{IEEEtran}

\usepackage{graphicx,amsmath,amssymb,verbatim,xspace}
\usepackage{color,subcaption,tabu,float,rotating,url,booktabs}
\usepackage[colorlinks=true,bookmarks=false]{hyperref}
\usepackage{epsfig}
\usepackage{graphicx}
\usepackage{amsmath}
\usepackage{amssymb}
\usepackage{bm}
\usepackage{wrapfig}
\usepackage{caption}
\usepackage{color}
\usepackage{xspace}

\newcommand{\COCO}{MS COCO\xspace}

\newcommand{\bleu}{BLEU\xspace}

\newcommand{\rouge}{ROUGE\xspace}

\newcommand{\cider}{CIDEr\xspace}
\newcommand{\mathcider}{\text{CIDEr}\xspace}
\newcommand{\mathciderD}{\text{CIDEr-D}\xspace}
\newcommand{\cidern}{CIDEr$_n$\xspace}

\newcommand{\meteor}{METEOR\xspace}
\newcommand{\rs}{s\xspace}
\newcommand{\RS}{S\xspace}
\newcommand{\ngram}{$n$-gram\xspace}
\newcommand{\ngrams}{$n$-grams\xspace}
\newcommand{\rougen}{ROUGE$_N$\xspace}
\newcommand{\rougel}{ROUGE$_L$\xspace}
\newcommand{\rouges}{ROUGE$_S$\xspace}
\newcommand{\capSm}{MS COCO c5\xspace}
\newcommand{\capLg}{MS COCO c40\xspace}

\begin{document}
\title{Microsoft COCO Captions: Data Collection and Evaluation Server}
\author{Xinlei Chen, Hao Fang, Tsung-Yi~Lin, Ramakrishna Vedantam \\ Saurabh Gupta, Piotr~Doll\'ar, C.~Lawrence~Zitnick
\IEEEcompsocitemizethanks{
\IEEEcompsocthanksitem Xinlei Chen is with Carnegie Mellon University.
\IEEEcompsocthanksitem Hao Fang is with the University of Washington.
\IEEEcompsocthanksitem T.Y.~Lin is with Cornell NYC Tech.
\IEEEcompsocthanksitem Ramakrishna Vedantam is with Virginia Tech.
\IEEEcompsocthanksitem Saurabh Gupta is with the Univeristy of California, Berkeley.
\IEEEcompsocthanksitem P.~Doll\'ar is with Facebook AI Research.
\IEEEcompsocthanksitem C.~L.~Zitnick is with Microsoft Research, Redmond.}
}

\IEEEcompsoctitleabstractindextext{\begin{abstract}
In this paper we describe the Microsoft COCO Caption dataset and evaluation
server. When completed, the dataset will contain over one and a half million captions describing over 330,000 images. For the training and validation images, five independent human generated captions will be provided. To ensure consistency in evaluation of automatic caption generation algorithms, an evaluation server is used. The evaluation server receives candidate captions and scores them using several popular metrics, including \bleu, \meteor, \rouge and \cider. Instructions for using the evaluation server are provided.  \end{abstract}} \maketitle

\section{Introduction}
The automatic generation of captions for images is a long standing and challenging problem in artificial intelligence ~\cite{barnard2001learning,barnard2003matching,lavrenko2003model,kulkarni2011baby,mitchell2012midge,farhadi2010every,hodosh2013framing,kuznetsova2012collective,yang2011corpus,gupta2012choosing,bruni2012distributional,feng2013automatic,elliott2013image,karpathy2014deep,gong2014improving,mason2014non,kuznetsova2014tree,RamnathBVESKHGYRBT14,lazaridou2014wampimuk}. Research in this area spans numerous domains, such as computer vision, natural language processing, and machine learning. Recently there has been a surprising resurgence of interest in this area \cite{ryan2014multimodal,mao2014explain,vinyals2014show,karpathy2014deep2,kiros2014unifying,donahue2014long,fang2014captions,chen2014learning,lebret2015phrase,lebret2014simple,lazaridou2015combining}, due to the renewed interest in neural network learning techniques \cite{Hinton,hochreiter1997long} and increasingly large datasets \cite{Imagenet,grubinger2006iapr,ordonez2011im2text,hodosh2013framing,young2014image,jianfu2015,COCO}.

In this paper, we describe our process of collecting captions for the Microsoft COCO Caption dataset, and the evaluation server we have set up to evaluate performance of different algorithms. The \COCO caption dataset contains human generated captions for images contained in the Microsoft Common Objects in COntext (COCO) dataset \cite{COCO}. Similar to previous datasets \cite{hodosh2013framing,young2014image}, we collect our captions using Amazon's Mechanical Turk (AMT). Upon completion of the dataset it will contain over a million captions.

When evaluating image caption generation algorithms, it is essential that a consistent evaluation protocol is used. Comparing results from different approaches can be difficult since numerous evaluation metrics exist \cite{bleu,rouge,meteor,cider}. To further complicate matters the implementations of these metrics often differ. To help alleviate these issues, we have built an evaluation server to enable consistency in evaluation of different caption generation approaches. Using the testing data, our evaluation server evaluates captions output by different approaches using numerous automatic metrics: \bleu \cite{bleu}, \meteor \cite{meteor}, \rouge \cite{rouge} and \cider \cite{cider}. We hope to augment these results with human evaluations on an annual basis.

This paper is organized as follows: First we describe the data collection process. Next, we describe the caption evaluation server and the various metrics used. Human performance using these metrics are provided. Finally the annotation format and instructions for using the evaluation server are described for those who wish to submit results. We conclude by discussing future directions and known issues.

\begin{figure}
  \centering
  \includegraphics[width=\linewidth]{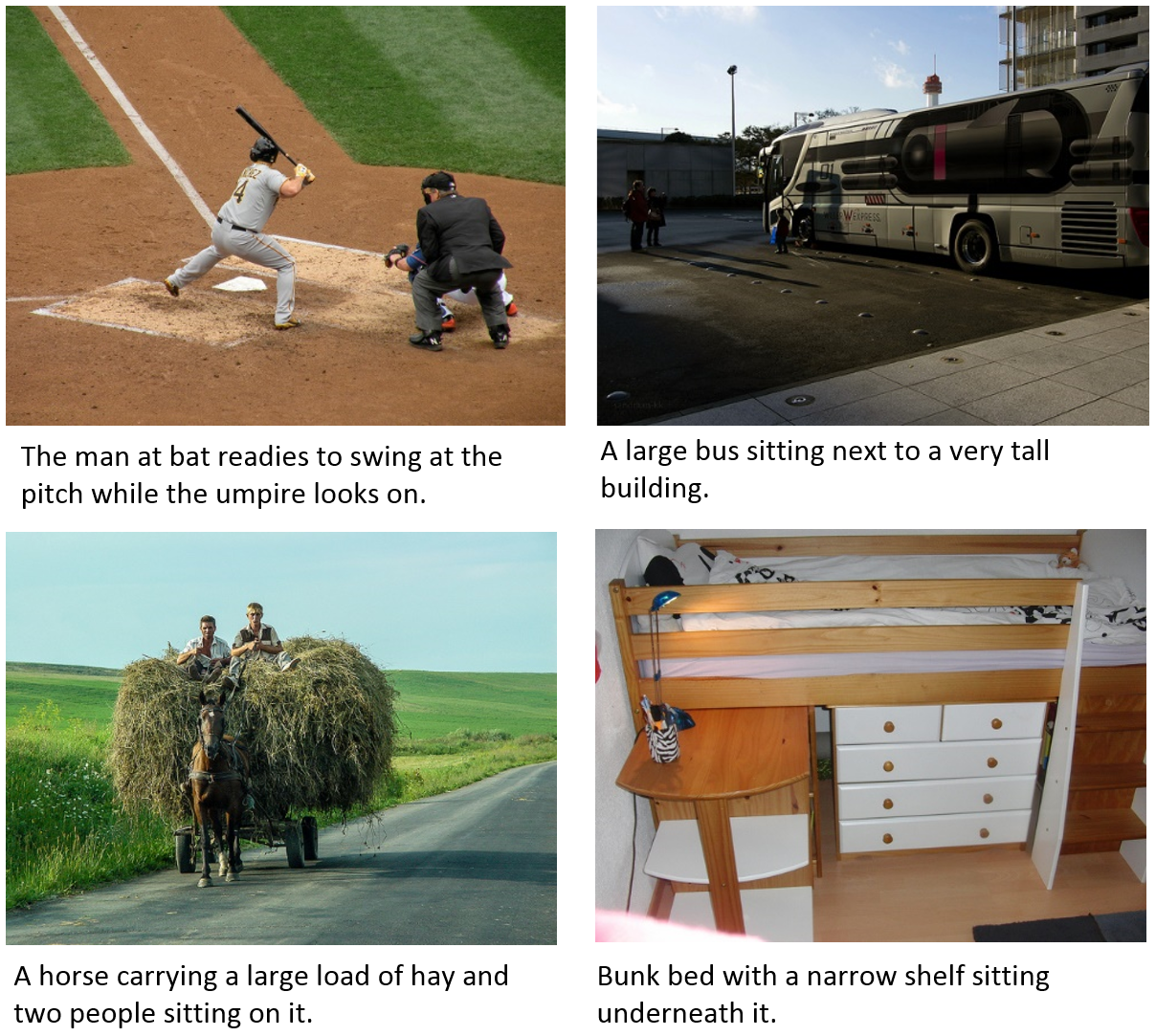}\\
  \caption{Example images and captions from the Microsoft COCO Caption dataset.}\label{fig:teaser}
\end{figure}

\section{Data Collection}

In this section we describe how the data is gathered for the MS COCO captions dataset. For images, we use the dataset collected by Microsoft COCO \cite{COCO}. These images are split into training, validation and testing sets. The images were gathered by searching for pairs of 80 object categories and various scene types on Flickr. The goal of the MS COCO image collection process was to gather images containing multiple objects in their natural context. Given the visual complexity of most images in the dataset, they pose an interesting and difficult challenge for image captioning.

For generating a dataset of image captions, the same training, validation and testing sets were used as in the original MS COCO dataset. Two datasets were collected. The first dataset \capSm contains five reference captions for every image in the \COCO training, validation and testing datasets. The second dataset \capLg contains 40 reference sentences for a randomly chosen 5,000 images from the \COCO testing dataset. \capLg was created since many automatic evaluation metrics achieve higher correlation with human judgement when given more reference sentences \cite{cider}. \capLg may be expanded to include the \COCO validation dataset in the future.

\begin{figure}
  \centering
  \includegraphics[width=\linewidth]{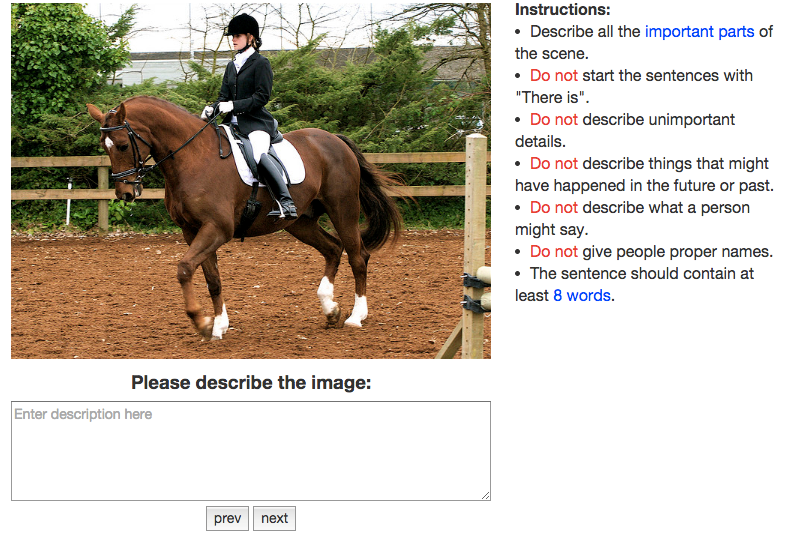}\\
  \caption{Example user interface for the caption gathering task.}\label{fig:UI}
\end{figure}

Our process for gathering captions received significant inspiration from the work of Young etal. \cite{young2014image} and Hodosh etal. \cite{hodosh2013framing} that collected captions on Flickr images using Amazon's Mechanical Turk (AMT). Each of our captions are also generated using human subjects on AMT. Each subject was shown the user interface in Figure \ref{fig:UI}. The subjects were instructed to:
\begin{itemize}\item Describe all the important parts of the scene.
\item Do not start the sentences with ``There is’’.
\item Do not describe unimportant details.
\item Do not describe things that might have happened in the future or past.
\item Do not describe what a person might say.
\item Do not give people proper names.
\item The sentences should contain at least 8 words.
\end{itemize}
The number of captions gathered is 413,915 captions for 82,783 images in training, 202,520 captions for 40,504 images in validation and 379,249 captions for 40,775 images in testing including 179,189 for \capSm and 200,060 for \capLg.  For each testing image, we collected one additional caption to compute the scores of human performance for comparing scores of machine generated captions.  The total number of collected captions is 1,026,459. We plan to collect captions for the MS COCO 2015 dataset when it is released, which should approximately double the size of the caption dataset. The AMT interface may be obtained from the \COCO website.

\section{Caption evaluation}
In this section we describe the \COCO caption evaluation server. Instructions for using the evaluation server are provided in Section \ref{instructions}. As input the evaluation server receives candidate captions for both the validation and testing datasets in the format specified in Section \ref{instructions}. The validation and test images are provided to the submitter. However, the human generated reference sentences are only provided for the validation set. The reference sentences for the testing set are kept private to reduce the risk of overfitting.

Numerous evaluation metrics are computed on both \capSm and \capLg. These include \bleu-1, \bleu-2, \bleu-3, \bleu-4, \rouge-L, \meteor and \cider-D. The details of the these metrics are described next.

\subsection{Tokenization and preprocessing}

Both the candidate captions and the reference captions are pre-processed by the
evaluation server.
To tokenize the captions, we use Stanford PTBTokenizer in Stanford CoreNLP tools
(version 3.4.1) \cite{corenlp} which mimics Penn Treebank 3 tokenization.
In addition, punctuations\footnote{The full list of punctuations:
\{``, '', `, ', -LRB-, -RRB-, -LCB-, -RCB-, ., ?, !, ,, :, -, --, ..., ;\}.} are
removed from the tokenized captions.

\subsection{Evaluation metrics}

Our goal is to automatically evaluate for an image $I_i$ the quality of a
candidate caption $c_i$ given a set of reference captions $\RS_{i} =
\{\rs_{i1},\ldots,\rs_{im}\} \in \RS$.
The caption sentences are represented using sets of \ngrams, where an
\ngram~$\omega_k \in \Omega$ is a set of one or more ordered words. In this
paper we explore \ngrams~with one to four words.
No stemming is performed on the words. The number of times an \ngram~$\omega_k$
occurs in a sentence $\rs_{ij}$ is denoted $h_k(\rs_{ij})$ or $h_k(c_i)$ for the
candidate sentence $c_i \in C$.

\subsection{\bleu}
\label{ss:bleu}
\bleu~\cite{bleu} is a popular machine translation metric that analyzes the
co-occurrences of \ngrams~between the candidate and reference sentences.
It computes a corpus-level clipped \ngram~precision between sentences as follows:
\begin{equation}
	CP_n(C , \RS) = \frac{\sum_{i} \sum_{k} \min ( h_k(c_i), \max\limits_{j \in m}h_k(\rs_{ij}))}
	{\sum_{i} \sum_{k} h_k(c_i)},
\end{equation}
where $k$ indexes the set of possible \ngrams\ of length $n$. The clipped
precision metric limits the number of times an \ngram~may be counted to the
maximum number of times it is observed in a single reference sentence. Note that
$CP_n$ is a precision score and it favors short sentences. So a brevity penalty
is also used:
\begin{equation}
b(C, \RS)= \begin{cases} 1 &\text{if } l_C > l_\RS \\
	      e^{1-l_\RS/l_C} &\text{if } l_C \le l_\RS
          \end{cases},
\end{equation}
where $l_C$ is the total length of candidate sentences $c_i$'s and $l_\RS$ is
the length of the corpus-level effective reference length.
When there are multiple references for a candidate sentence, we choose to use
the {\it closest} reference length for the brevity penalty.

The overall \bleu~score is computed using a weighted geometric mean of the
individual \ngram~precision:
\begin{align}
	\bleu_{N}(C, \RS) = b(C, \RS) \exp\left( \sum_{n=1}^N w_n \log CP_n(C, \RS)\right),
\end{align}
where $N = 1, 2, 3, 4$ and $w_n$ is typically held constant for all $n$.

\bleu~has shown good performance for corpus-level comparisons over which a high
number of \ngram~matches exist. However, at a sentence-level the \ngram~matches
for higher $n$ rarely occur. As a result, \bleu~performs poorly when comparing
individual sentences.
\subsection{ROUGE}
\label{ss:rouge}
ROUGE~\cite{rouge} is a set of evaluation metrics designed to evaluate text summarization algorithms.
\begin{enumerate}
\item \rougen:
The first \rouge~metric computes a simple \ngram~recall over all reference summaries given a candidate sentence:
\begin{equation}
ROUGE_N( c_i,\RS_{i}) = \frac{\sum_j \sum_k \min(h_k(c_i), h_k(\rs_{ij}))} {\sum_j \sum_k h_k(\rs_{ij})}
\end{equation}

\item \rougel:
\rougel~ uses a measure based on the Longest Common Subsequence (LCS). An LCS is a set words shared by two sentences which occur in the same order. However, unlike \ngrams~there may be words in between the words that create the LCS. Given the length $l(c_i,\rs_{ij})$ of the LCS between a pair of sentences, \rougel~is found by computing an F-measure:

\begin{align} R_{l} &= \max\limits_{j}\frac{l(c_i,\rs_{ij})}{|\rs_{ij}|} \\
		P_{l} &= \max\limits_{j}\frac{l(c_i,\rs_{ij})}{|c_i|} \\
		ROUGE_L( c_i,\RS_{i}) &= \frac{(1 + \beta^2) R_{l} P_{l}}{R_{l} + \beta^2 P_{l}}
\end{align}
$R_l$ and $P_l$ are recall and precision of LCS. $\beta$ is usually set to favor \emph{recall} ($\beta = 1.2$). Since \ngrams~are implicit in this measure due to the use of the LCS, they need not be specified.

\item \rouges:
The final \rouge~metric uses skip bi-grams instead of the LCS or \ngrams. Skip bi-grams are pairs of ordered words in a sentence. However, similar to the LCS, words may be skipped between pairs of words. Thus, a sentence with 4 words would have $C^{4}_{2} = 6$ skip bi-grams. Precision and recall are again incorporated to compute an F-measure score. If $f_k(s_{ij})$ is the skip bi-gram count for sentence $s_{ij}$, \rouges~is computed as:
\begin{align}
R_{s} &= \max\limits_{j} \frac{\sum_k \min(f_k(c_i), f_k(\rs_{ij}))}{\sum_k f_k(\rs_{ij})} \\
P_{s} &= \max\limits_{j} \frac{\sum_k \min (f_k(c_i), f_k(\rs_{ij}))}{\sum_k f_k(c_{i})} \\
ROUGE_S( c_i,\RS_{i}) &= \frac{(1 + \beta^2) R_{s} P_{s}}{R_{s} + \beta^2 P_{s}}
\end{align}
Skip bi-grams are capable of capturing long range sentence structure. In practice, skip bi-grams are computed so that the component words occur at a distance of at most 4 from each other.
\end{enumerate}
\subsection{\meteor}
\meteor~\cite{meteor} is calculated by generating an alignment between the words in the candidate and reference sentences, with an aim of 1:1 correspondence. This alignment is computed while minimizing the number of chunks, $ch$, of contiguous and identically ordered tokens in the sentence pair. The alignment is based on exact token matching, followed by WordNet synonyms~\cite{miller1995wordnet}, stemmed tokens and then paraphrases. Given a set of alignments, $m$, the \meteor score is the harmonic mean of precision $P_m$ and recall $R_m$ between the best scoring reference and candidate:
\begin{align}
Pen = \gamma \left( \frac{ch}{m} \right)^\theta \\
F_{mean} = \frac{P_m R_m}{\alpha P_m + (1-\alpha) R_m} \\
P_m = \frac{|m|}{\sum_{k} h_k(c_i)}\\
R_m = \frac{|m|}{\sum_{k} h_k(s_{ij})}\\
METEOR = (1-Pen) F_{mean}
\end{align}
Thus, the final \meteor score includes a penalty $Pen$ based on chunkiness of resolved matches and a harmonic mean term that gives the quality of the resolved matches. The default parameters $\alpha$, $\gamma$ and $\theta$ are used for this evaluation. Note that similar to \bleu, statistics of precision and recall are first aggregated over the entire corpus, which are then combined to give the corpus-level \meteor score.

\subsection{\cider}
The CIDEr metric~\cite{cider} measures consensus in image captions by performing a Term Frequency Inverse Document Frequency (TF-IDF) weighting for each \ngram. The number of times an \ngram~$\omega_k$ occurs in a reference sentence $\rs_{ij}$ is denoted by $h_k(\rs_{ij})$ or $h_k(c_i)$ for the candidate sentence $c_i$. CIDEr computes the TF-IDF weighting $g_k(\rs_{ij})$ for each \ngram~$\omega_k$ using:

\begin{multline}
g_k(\rs_{ij}) = \\ \frac{h_k(\rs_{ij})}{\sum_{\omega_l \in \Omega} h_l(\rs_{ij})} \log\left(\frac{|I|}{\sum_{I_p \in I} \min(1, \sum_q h_k(s_{pq}))}\right),
\end{multline}

where $\Omega$ is the vocabulary of all \ngrams and $I$ is the set of all images in the dataset. The first term measures the TF of each \ngram~$\omega_k$, and the second term measures the rarity of $\omega_k$ using its IDF.  Intuitively, TF places higher weight on \ngrams~that frequently occur in the reference sentences describing an image, while IDF reduces the weight of \ngrams~that commonly occur across all descriptions. That is, the IDF provides a measure of word saliency by discounting popular words that are likely to be less visually informative. The IDF is computed using the logarithm of the number of images in the dataset $|I|$ divided by the number of images for which $\omega_k$ occurs in any of its reference sentences.

The \cidern~score for \ngrams~of length $n$ is computed using the average cosine similarity between the candidate sentence and the reference sentences, which accounts for both precision and recall:
\begin{equation}
\mathcider_n(c_i, \RS_i) = \frac{1}{m}\sum_j \frac{\bm{g^n}(c_{i})\cdot \bm{g^n}(\rs_{ij})}{\|\bm{g^n}(c_{i})\|\|\bm{g^n}(\rs_{ij})\|},
\end{equation}
where $\bm{g^n}(c_{i})$ is a vector formed by $g_k(c_{i})$ corresponding to all \ngrams\ of length $n$ and $\|\bm{g^n}(c_{i})\|$ is the magnitude of the vector $\bm{g^n}(c_{i})$. Similarly for $\bm{g^n}(\rs_{ij})$.

Higher order (longer) \ngrams~to are used to capture grammatical properties as well as richer semantics. Scores from \ngrams~of varying lengths are combined as follows:

\begin{equation} \label{eq:cider}
\mathcider(c_i, \RS_i) = \sum_{n=1}^N w_n \mathcider_n(c_i, \RS_i),
\end{equation}
Uniform weights are used $w_n=1/N$. $N$ = 4 is used.

\textbf{CIDEr-D} is a modification to \cider to make it more robust to gaming. Gaming refers to the phenomenon where a sentence that is poorly judged by humans tends to score highly with an automated metric. To defend the \cider metric against gaming effects, ~\cite{cider} add clipping and a length based gaussian penalty to the \cider metric described above. This results in the following equations for CIDEr-D:

\begin{multline}
\mathciderD_n(c_i, \RS_i) = \frac{10}{m}\sum_j e^{\frac{-(l(c_{i})-l(\rs_{ij}))^2}{2\sigma^2}} * \\ \frac{min(\bm{g^n}(c_{i}),\bm{g^n}(\rs_{ij})) \cdot \bm{g^n}(\rs_{ij})}{\|\bm{g^n}(c_{i})\|\|\bm{g^n}(\rs_{ij})\|},
\end{multline}

Where $l(c_{i})$ and $l(\rs_{ij})$ denote the lengths of candidate and reference sentences respectively. $\sigma = 6$ is used. A factor of 10 is used in the numerator to make the CIDEr-D scores numerically similar to the other metrics.

The final CIDEr-D metric is computed in a similar manner to CIDEr (analogous to eqn. \ref{eq:cider}):
\begin{equation}
\mathciderD(c_i, \RS_i) = \sum_{n=1}^N w_n \mathciderD_n(c_i, \RS_i),
\end{equation}
Note that just like the \bleu and \rouge metrics, CIDEr-D does not use stemming. We adopt the CIDEr-D metric for the evaluation server.

\section{Human performance}
In this section, we study the human agreement among humans at this task. We start with analyzing the inter-human agreement for image captioning (Section.~\ref{sec:human-agreement-caption}) and then analyze human agreement for the word prediction sub-task and provide a simple model which explains human agreement for this sub-task (Section.~\ref{sec:human-agreement-word}).

\subsection{Human Agreement for Image Captioning}\label{sec:human-agreement-caption}
When examining human agreement on captions, it becomes clear that there are many equivalent ways to say essentially the same thing. We quantify this by conducting the following experiment: We collect one additional human caption for each image in the test set and treat this caption as the prediction. Using the \COCO caption evaluation server we compute the various metrics. The results are tabulated in Table \ref{tab:human-agreement-caption}.

\begin{table}\centering\footnotesize 
\caption{Human Agreement for Image Captioning: Various metrics when benchmarking a human generated caption against ground truth captions.}
\label{tab:human-agreement-caption}\vspace{2mm}
\begin{tabular}{lcc}
\toprule
 Metric Name & \capSm & \capLg \\
\midrule
\bleu 1      & 0.663 & 0.880 \\
\bleu 2      & 0.469  & 0.744 \\
\bleu 3      & 0.321 & 0.603\\
\bleu 4      & 0.217 & 0.471 \\
\midrule
\meteor     & 0.252 & 0.335 \\
\rougel      & 0.484 & 0.626 \\
\cider-D      & 0.854 & 0.910 \\
\bottomrule
\end{tabular}
\end{table}

\subsection{Human Agreement for Word Prediction}\label{sec:human-agreement-word}
We can do a similar analysis for human agreement at the sub-task of word prediction. Consider the task of tagging the image with words that occur in the captions. For this task, we can compute the human precision and recall for a given word $w$ by benchmarking words used in the $k+1$ human caption with respect to words used in the first $k$ reference captions. Note that we use weighted versions of precision and recall, where each negative image has a weight of 1 and each positive image has a weight equal to the number of captions containing the word $w$. Human precision ($H_p$) and human recall ($H_r$) can be computed from the counts of how many subjects out of $k$ use the word $w$ to describe a given image over the whole dataset.

\begin{figure*}\centering
 \includegraphics[width=0.22\textwidth]{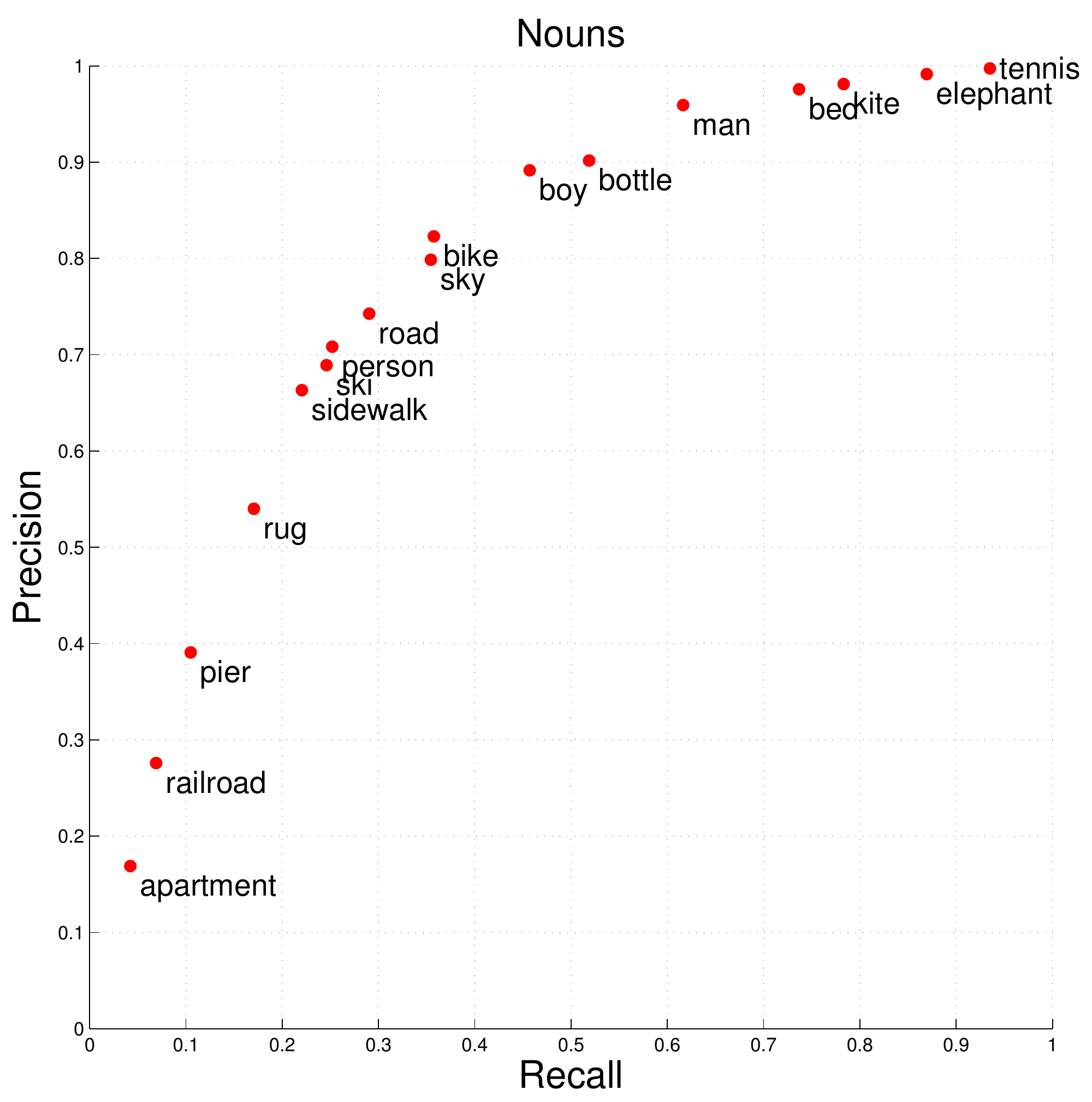}
 \includegraphics[width=0.22\textwidth]{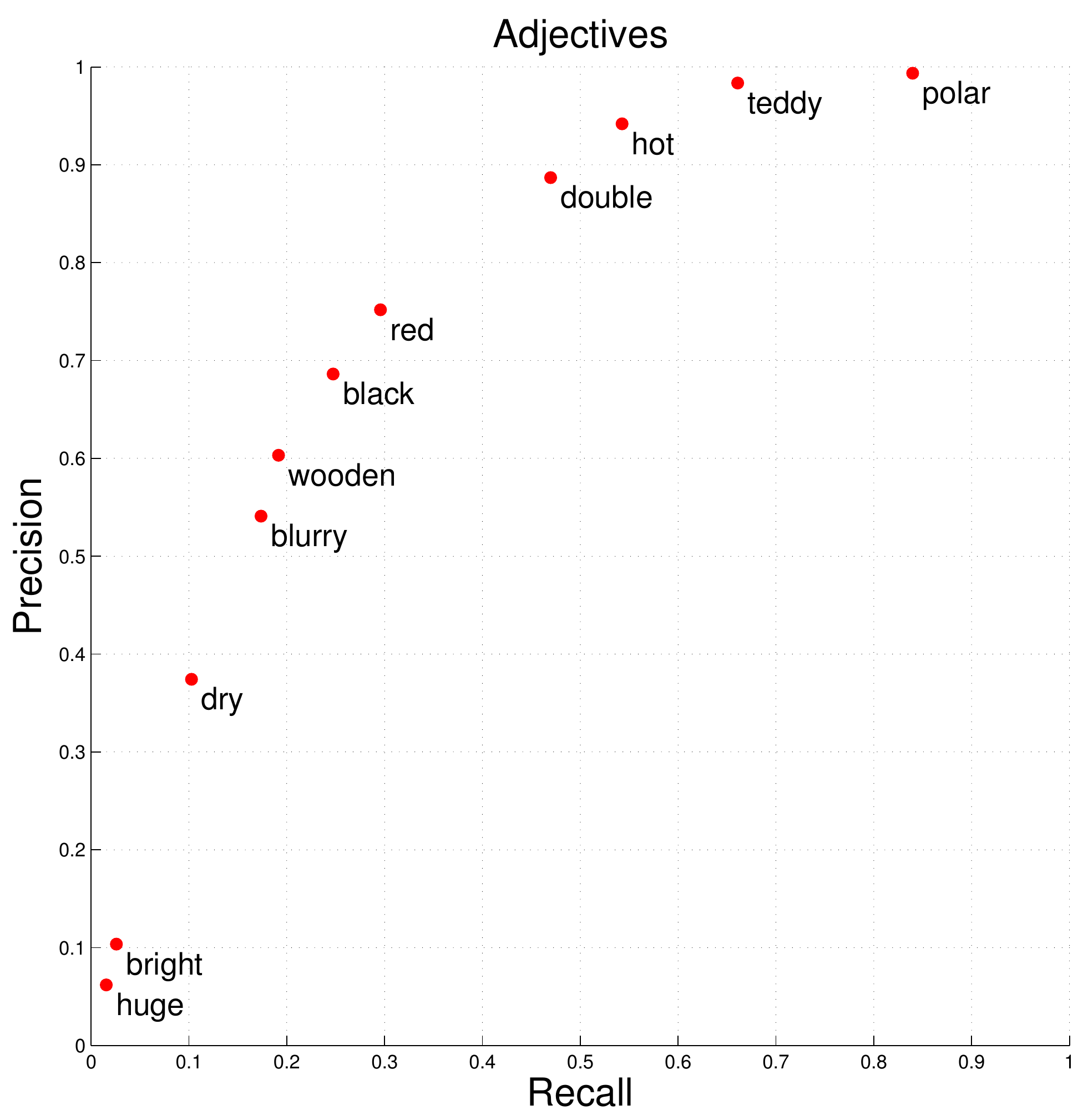}
 \includegraphics[width=0.22\textwidth]{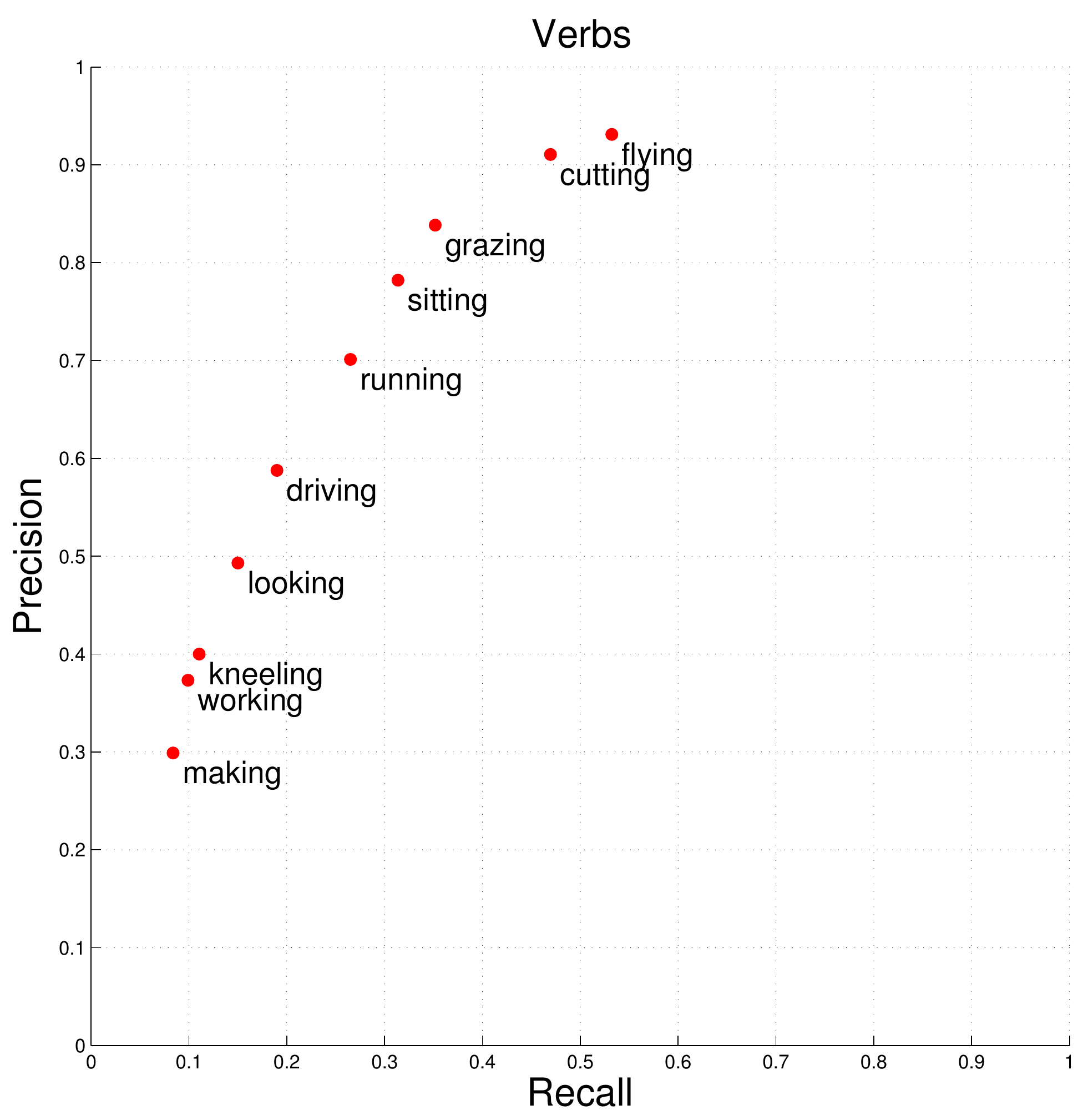} \\
 \includegraphics[width=0.34\textwidth]{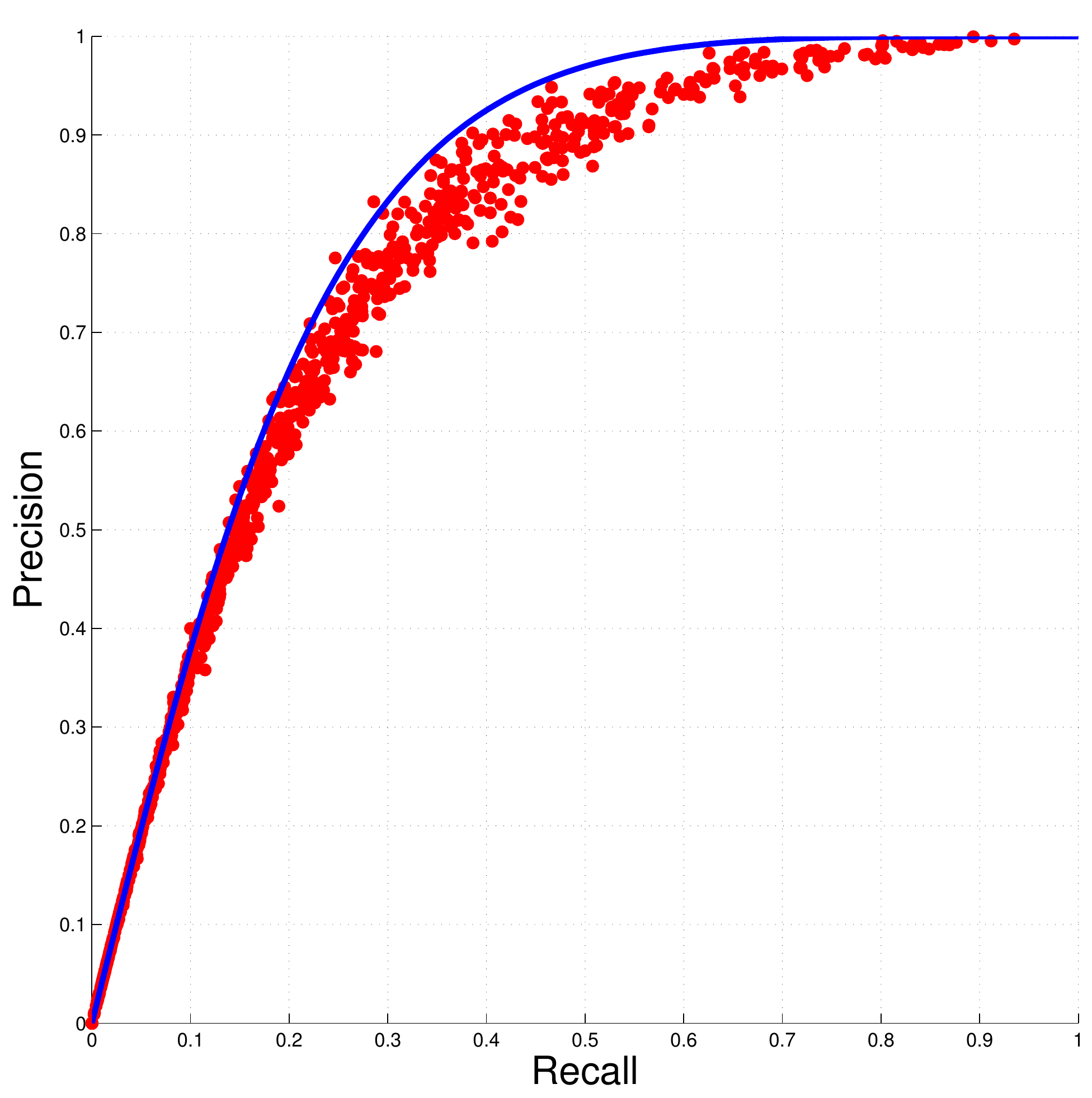}
 \includegraphics[width=0.34\textwidth]{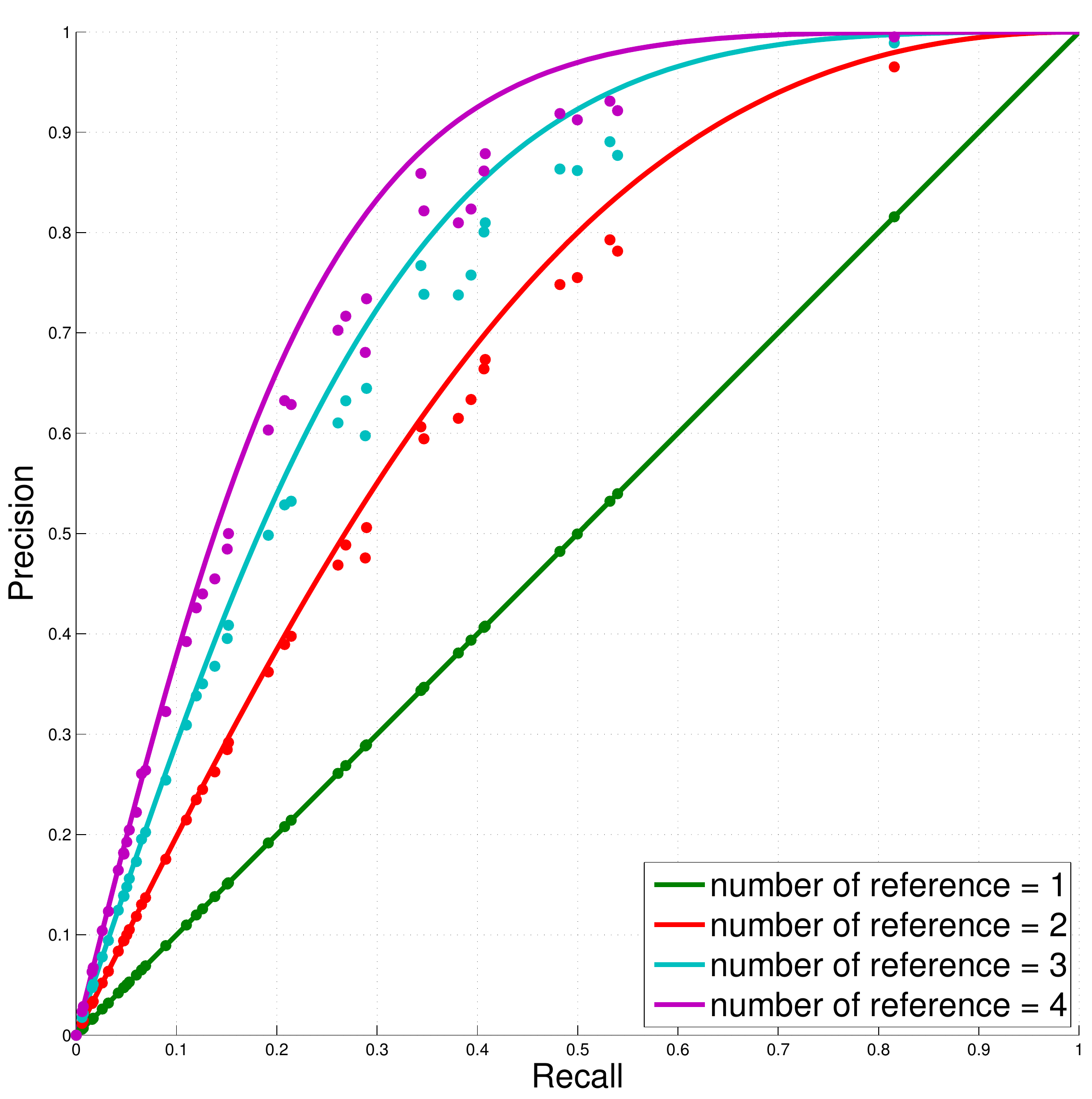}
\caption{Precision-recall points for human agreement: we compute precision and recall by treating one human caption as prediction and benchmark it against the others to obtain points on the precision recall curve. We plot these points for example nouns (top left), adjectives (top center), and verbs (top right), and for all words (bottom left). We also plot the fit of our model for human agreement with the empirical data (bottom left) and show how the human agreement changes with different number of captions being used (bottom right). We see that the human agreement point remains at the same recall value but dips in precision when using fewer captions.}
\label{fig:hpr}
\end{figure*}

We plot $H_p$ versus $H_r$ for a set of nouns, verbs and adjectives, and all 1000 words considered in Figure~\ref{fig:hpr}. Nouns referring to animals like `elephant' have a high recall, which means that if an `elephant' exists in the image, a subject is likely to talk about it (which makes intuitive sense, given `elephant' images are somewhat rare, and there are no alternative words that could be used instead of `elephant'). On the other hand, an adjective like `bright' is used inconsistently and hence has low recall. Interestingly, words with high recall also have high precision. Indeed, all the points of human agreement appear to lie on a one-dimensional curve in the two-dimension precision-recall space.

This observation motivates us to propose a simple model for when subjects use a particular word $w$ for describing an image. Let $o$ denote an object or visual concept associated with word $w$, $n$ be the total number of images, and $k$ be the number of reference captions. Next, let $q=P(o=1)$ be the probability that object $o$ exists in an image. For clarity these definitions are summarized in Table \ref{tab:human-defines}. We make two simplifications. First, we ignore \emph{image level saliency} and instead focus on \emph{word level saliency}. Specifically, we only model $p=P(w=1|o=1)$, the probability a subject uses $w$ given that $o$ is in the image, without conditioning on the image itself. Second, we assume that $P(w=1 | o=0) = 0$, i.e.~that a subject does not use $w$ unless $o$ is in the image. As we will show, even with these simplifications our model suffices to explain the empirical observations in Figure \ref{fig:hpr} to a reasonable degree of accuracy.

\begin{table}\centering\footnotesize
\caption{Model defintions.}
\label{tab:human-defines}\vspace{1mm}
\begin{tabular}{rcl}
\toprule
 $o$ & = & object or visual concept \\
 $w$ & = & word associated with $o$ \\
 $n$ & = & total number of images  \\
 $k$ & = & number of captions per image \\
 $q$ & = & $P(o=1)$ \\
 $p$ & = & $P(w = 1 | o = 1) $\\
\bottomrule
\end{tabular}\vspace{-2mm}
\end{table}

Given these assumptions, we can model human precision $\widetilde{H_p}$ and recall $\widetilde{H_r}$ for a word $w$ given only $p$ and $k$. First, given $k$ captions per image, we need to compute the expected number of (1) captions containing $w$ ($cw$), (2) true positives ($tp$), and (3) false positives ($fp$). Note that in our definition there can be up to $k$ true positives per image (if $cw=k$, i.e.~each of the $k$ captions contains word $w$) but at most 1 false positive (if none of the $k$ captions contains $w$). The expectations, in terms of $k$, $p$, and $q$ are:
\begin{eqnarray*}
E[cw] &=& \Sigma_{i=1}^k P(w^i=1) \\
      &=& \Sigma_i P(w^i = 1|o = 1)P(o = 1) \\
      & & + \Sigma_i P(w^i = 1|o = 0)P(o = 0) \\
      &=& k p q + 0 = \boxed{kpq} \\
E[tp] &=& \Sigma_{i=1}^{k} P(w^i = 1 \wedge w^{k+1} = 1) \\
      &=& \Sigma_i P(w^i = 1 \wedge w^{k+1} = 1 | o=1)P(o = 1) \\
      & & + \Sigma_i P(w^i = 1 \wedge w^{k+1} = 1 | o=0)P(o = 0) \\
      &=& kppq + 0= \boxed{kp^2q} \\
E[fp] &=& P(w^{1} \ldots w^{k} = 0 \wedge w^{k+1} = 1) \\
      &=& P(o = 1 \wedge w^{1} \ldots w^{k} = 0 \wedge w^{k+1} = 1) \\
      & & + P(o = 0 \wedge w^{1} \ldots w^{k} = 0 \wedge w^{k+1} = 1) \\
      &=& q (1-p)^{k} p +0 = \boxed{q(1-p)^{k}p}
\end{eqnarray*}
In the above $w^i=1$ denotes that $w$ appeared in the $i^{th}$ caption. Note that we are also assuming independence between subjects conditioned on $o$. We can now define model precision and recall as:
\begin{eqnarray*}
\widetilde{H_p} &:=& \frac{nE[tp]} {nE[tp] + nE[fp]}
                  = \frac{pk}{pk + (1-p)^{k}} \\
\widetilde{H_r} &:=& \frac{nE[tp]} {nE[cw]} = p
\end{eqnarray*}
Note that these expressions are independent of $q$ and only depend on $p$. Interestingly, because of the use of weighted precision and recall, the recall for a category comes out to be exactly equal to $p$, the probability a subject uses $w$ given that $o$ is in the image.

We set $k=4$ and vary $p$ to plot $\widetilde{H_p}$ versus $\widetilde{H_r}$, getting the curve as shown in blue in Figure~\ref{fig:hpr}~(bottom left). The curve explains the observed data quite well, closely matching the precision-recall tradeoffs of the empirical data (although not perfectly). We can also reduce the number of captions from four, and look at how the empirical and predicted precision and recall change. Figure~\ref{fig:hpr}~(bottom right), shows this variation as we reduce the number of reference captions per image from four to one annotations. We see that the points of human agreement remain at the same recall value, but decrease in their precision, which is consistent with what the model predicts. Also, the human precision at infinite subjects will approach one, which is again reasonable given that a subject will only use the word $w$ if the corresponding object is in the image (and in the presence of infinite subjects someone else will also use the word $w$).

In fact, the fixed recall value can help us recover $p$, the probability that a subject will use the word $w$ in describing the image given the object is present. Nouns like `elephant' and `tennis' have large $p$, which is reasonable. Verbs and adjectives, on the other hand, have smaller $p$ values, which can be justified from the fact that a) subjects are less likely to describe attributes of objects and b) subjects might use a different word (synonym) to describe the same attribute.

This analysis of human agreement also motivates using a different metric for measuring performance. We propose Precision at Human Recall (PHR) as a metric for measuring performance of a vision system performing this task. Given that human recall for a particular word is fixed and precision varies with the number of annotations, we can look at system precision at human recall and compare it with human precision to report the performance of the vision system.

\section{Evaluation Server Instructions}
\label{instructions}

Directions on how to use the \COCO caption evaluation server can be found on the \href{http://mscoco.org/dataset/#upload}{\COCO website}. The evaluation server is hosted by \href{https://www.codalab.org/competitions/3221}{CodaLab}. To participate, a user account on CodaLab must be created. The participants need to generate results on both the validation and testing datasets. When training for the generation of results on the test dataset, the training and validation dataset may be used as the participant sees fit. That is, the validation dataset may be used for training if desired. However, when generating results on the validation set, we ask participants to only train on the training dataset, and only use the validation dataset for tuning meta-parameters. Two JSON files should be created corresponding to results on each dataset in the following format:
$$\begin{array}{lcl}

[\{ & & \\
``image\_id" & : & int, \\
``caption" & : & str, \\
\}] &  &\\

\end{array}$$
The results may then be placed into a zip file and uploaded to the server for evaluation. Code is also provided on \href{https://github.com/tylin/coco-caption}{GitHub} to evaluate results on the validation dataset without having to upload to the server. The number of submissions per user is limited to a fixed amount.

\section{Discussion}

Many challenges exist when creating an image caption dataset. As stated in \cite{hodosh2013framing, cider,elliott2014comparing} the captions generated by human subjects can vary significantly. However even though two captions may be very different, they may be judged equally ``good'' by human subjects. Designing effective automatic evaluation metrics that are highly correlated with human judgment remains a difficult challenge \cite{hodosh2013framing,cider,elliott2014comparing,callison2006re}. We hope that by releasing results on the validation data, we can help enable future research in this area.

Since automatic evaluation metrics do not always correspond to human judgment, we hope to conduct experiments using human subjects to judge the quality of automatically generated captions, which are most similar to human captions, and whether they are grammatically correct \cite{elliott2014comparing,cider,hodosh2013framing,kulkarni2011baby,mitchell2012midge}. This is essential to determining whether future algorithms are indeed improving, or whether they are merely over fitting to a specific metric. These human experiments will also allow us to evaluate the automatic evaluation metrics themselves, and see which ones are correlated to human judgment.

\bibliographystyle{IEEEtran}
\bibliography{coco}


\end{document}